\documentclass[11pt]{article}
\usepackage{acl2015}
\usepackage{times}
\usepackage{url}
\usepackage{latexsym}

\newif\ifMOVIE
\MOVIEfalse

\usepackage{ascmac}
\usepackage{graphicx}

\usepackage{algorithm}
\usepackage{algorithmic}

\title{A Gamification of Japanese Dependency Parsing}
\author{Masayuki Asahara}
\date{}

\begin{document}

\maketitle

\begin{abstract}
Gamification approaches have been used as a way for creating language resources
 for NLP. It is also used for presenting and teaching the algorithms in NLP and linguistic phenomena.
This paper argues about a design of gamification for Japanese syntactic
 dependendency parsing for the latter objective.
The user interface design is based on a transition-based shift reduce dependency
 parsing which needs only two actions of SHIFT (not attach) and REDUCE
 (attach) in Japanese dependency structure.
We assign the two actions for two-way directional control on a gamepad or other devices.
We also design the target sentences from psycholinguistics researches.
\end{abstract}

\section{Introduction}
We start this research with a naive question: how to make the syntactic dependency annotation more simple and joyful for non-specialist by associating the game devices.  Though gold standard annotation might be developped by the specialist, we would like to collect the distribution of the judges by non-specialists in order to know which sentences are hard to parse for human.

We propose a novel game application to test human dependency parsing process.
The game is named ``shWiiFit Reduce Dependency Parsing'' of which movieclip can be viewed by clicking Figure \ref{swfrdp} or browsing the URL \url{http://goo.gl/cWncIi}.
The idea is that a player stands on the balance board, reads a sentence to parse and selects one of two possible actions (``SHIFT'' or ``REDUCE'') by moving one's weight to the left or right depending on whether a pair of focused phrases are in syntactic dependency relation or not.

In this article, we focus on Japanese syntactic dependency parsing.
Reasons are: 1) Japanese syntactic dependency relations are simpler to parse due
to the strictly head final characteristics. 
2) State-of-the-art parsers for Japanese are mostly transition-based \cite{Kudo02,Iwatate08}.
3) There is a linear-order, transition-based, and above all, an $O(N)$ algorithm \cite{Sassano04} which is a simplified variation of \cite{Nivre04}'s shift reduce-like algorithm.

Section \ref{sec:swfrdp} shows the basic design of game user interface.
Section \ref{sec:gps} presents design of sample sentences based on psycholinguistic researches.
Section \ref{sec:experiments} reports experiments and the analysis of the results.
Section \ref{sec:conclustions} states conclusions and our future work.

\ifMOVIE
\begin{figure}[htb]
\begin{center}
 \includemovie[toolbar,mouse,
	text={\begin{minipage}[t]{7cm}
 \begin{center}
\includegraphics[width=7cm]{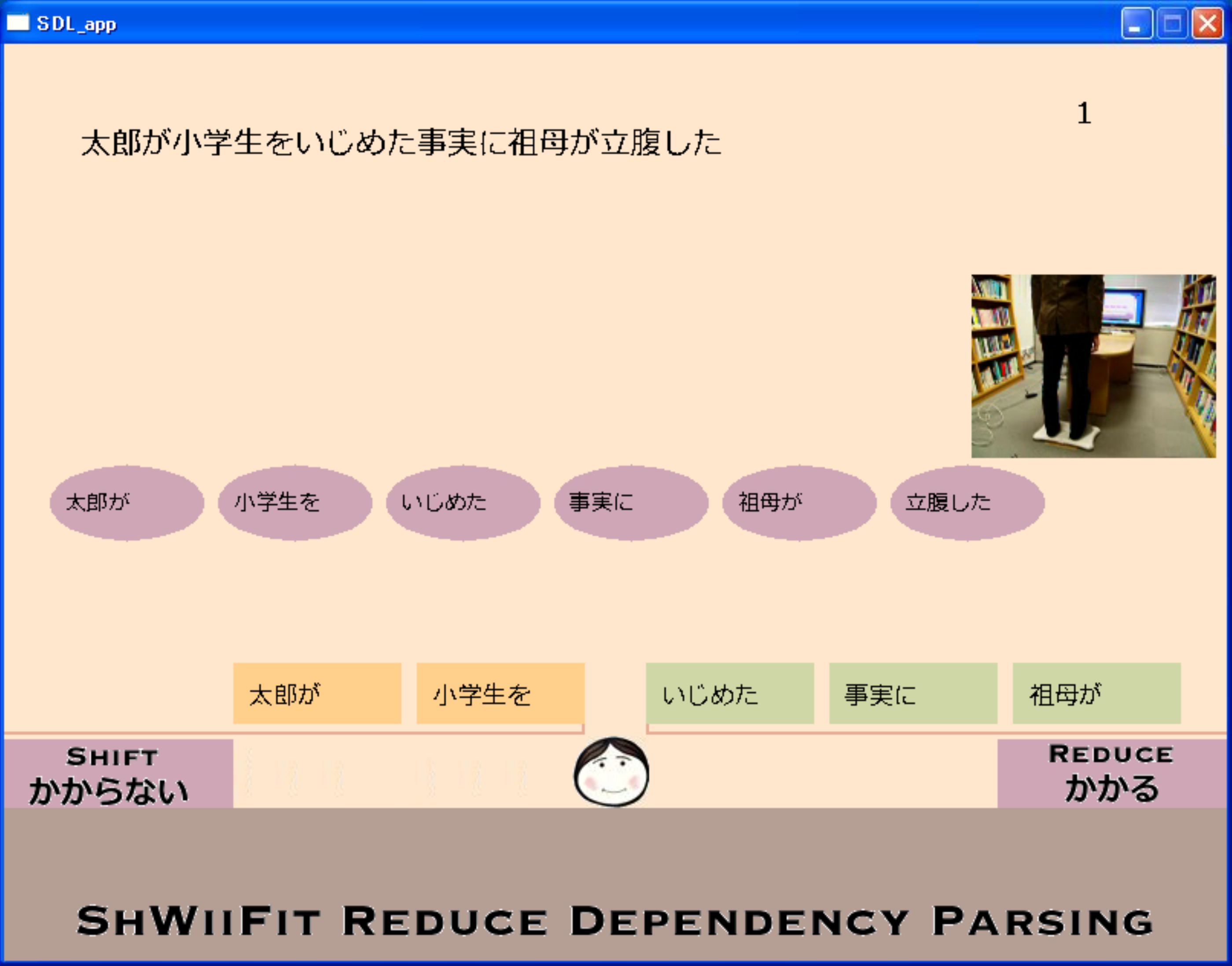}
\url{http://goo.gl/cWncIi}\\
\end{center}
 \end{minipage}
 },
 ]{}{}{fig/swfrdp.mp4}
 \end{center}
\caption{shWiiFit Reduce Dependency Parsing \label{swfrdp}}
\end{figure}
\else
\begin{figure}[htb]
\begin{center}
\includegraphics[width=7cm]{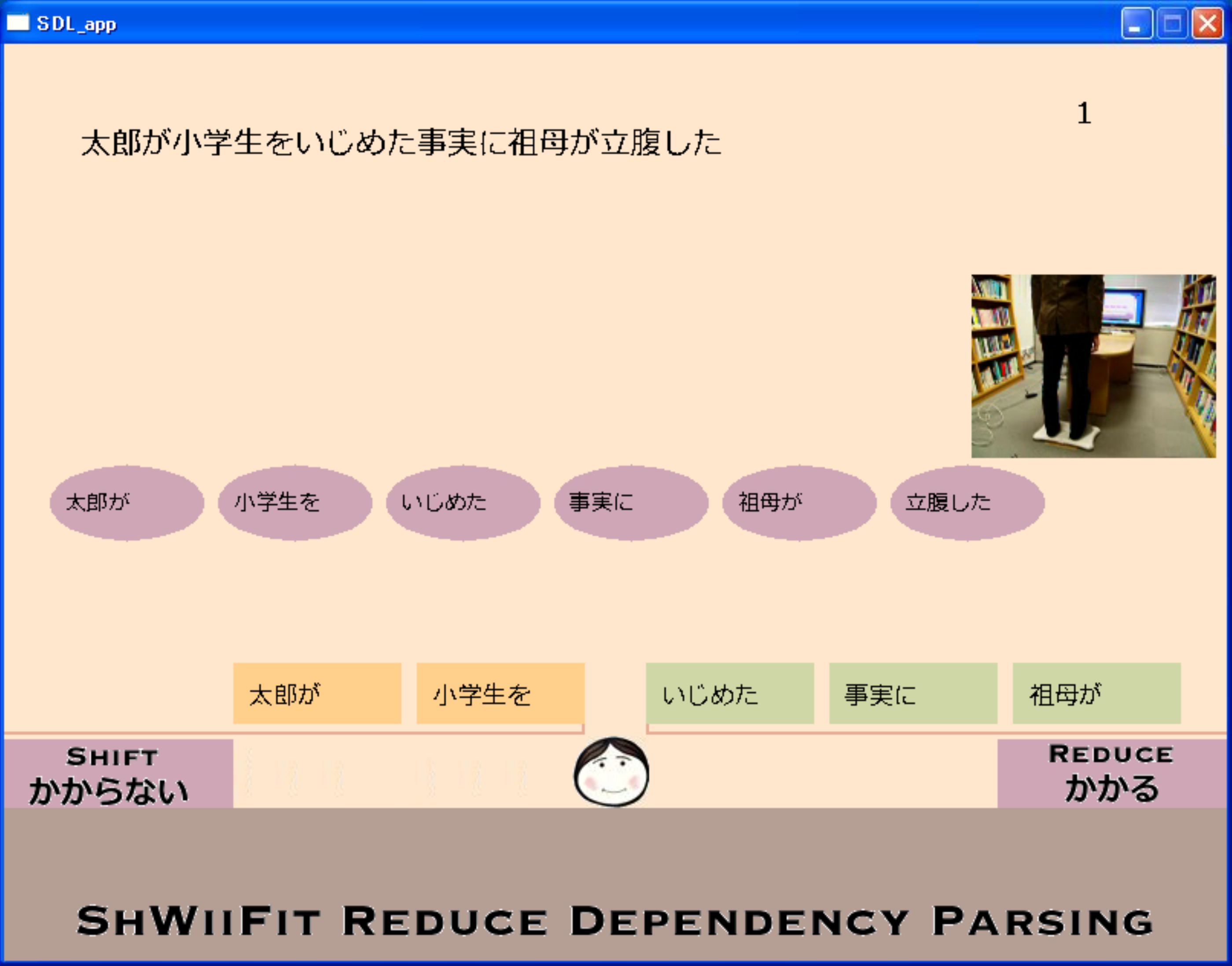}
\url{http://goo.gl/cWncIi}\\
\end{center}
\caption{shWiiFit Reduce Dependency Parsing \label{swfrdp}}
\end{figure}
\fi

\section{User Interface Design of Gamification \label{sec:swfrdp}}
\begin{algorithm}[tb]
\caption{Shift reduce-like Japanese dependency parsing}
\begin{algorithmic}
\STATE \% Initialization
\STATE $\langle S, Q, A \rangle = \langle nil, W, \phi \rangle$
\REPEAT
\IF{$S == nil$}
\STATE \% ``default shift''
\STATE $\langle nil, q|Q, A \rangle \Rightarrow \langle q, Q, A \rangle$
\ELSIF{$|Q| == 1$} 
\STATE \% ``default reduce''
\STATE $\langle s|S, q, A \rangle \Rightarrow \langle S, q, A \cup (s,q) \rangle$
\ELSE
\STATE \% Judge on $\langle s|S, q|Q, A \rangle$
\IF{$s$ and $q$ has a dependency relation}
\STATE \% ``REDUCE''
\STATE $\langle s|S, q|Q, A \rangle \Rightarrow \langle S, q|Q, A \cup(s,q) \rangle$
\ELSE
\STATE \% ``SHIFT''
\STATE $\langle s|S, q|Q, A \rangle \Rightarrow \langle q|s|S, Q, A \rangle$
\ENDIF
\ENDIF
\UNTIL{$S == nil$ \AND $|Q| == 1$}
\RETURN $A$
\end{algorithmic}
\end{algorithm}

We describe a fitness game application called ``shWiiFit Reduce Dependency Parsing''.
The screen is shown in Figure \ref{swfrdp}.
The game used Sassano's shift reduce-like Algorithm (Algorithm 1) which parses a length-$N$ sentence at most $2N$ actions.
``SHIFT'' and ``REDUCE'' are the actions which require human (or machine) judgements on dependency relations.

We introduce a tuple of stack $S$, queue $Q$, and the set of dependency relations $A$, initializing nil, an input phrase sequence, and empty respectively.
A parsing game proceeds by state transitions of the tuple as the following actions.
When $S$ is nil (and there are multiple phrases in $Q$), the game automatically executes the action ``default shift''.
The first element $q$ of $Q$ is moved onto $S$.
When $S$ is not nil and $Q$ has only single phrase $q$ left, the game automatically executes action ``default reduce''.
As the top element $s$ of $S$ is by default a dependent of $q$,
the pair $\langle s, q \rangle$ is added to $A$ and $s$ is removed from $S$.
When the top element $s$ of $S$ and the first element $q$ of $Q$ are both defined, the player judges whether $q$ is the head of $s$ and inputs an answer.
If the answer is yes, then the game executes the action ``REDUCE'' and the pair  $\langle s, q \rangle$ is added to $A$ and $s$ is removed from $S$.
Otherwise, the game executes the action ``SHIFT'' and the first element $q$ is moved on the top of $S$.
The game ends when $S$ is nil and there is only the last phrase of an input sentence in $Q$.

A player stands on a balance board and watches a screen as shown in Figure \ref{swfrdp} during a game.
A sentence to parse is displayed in the top of the screen.
Below it, dependency relations that a player builts during a game is shown accordingly.

At the start of a game, the face icon in the bottom is positioned center.
No phrase is in stack $S$ shown in the left hand side of the face icon, while all 6 phrases are in queue $Q$ shown in the right hand side of the face icon.
During the game, a player judges whether the first (leftmost) phrase $q$ of queue $Q$ is the head of the top (rightmost) phrase $s$ of stack $S$.
If so, the player weighs to the right so as for the face icon to move towards the REDUCE wall.
Otherwise, the player weighs to the left so as for the face icon to move towards the SHIFT wall.
The game shows 820-860 milliseconds animations of screen transitions just after executing each action, to move the icon automatically centered back.
At the end of the game, the screen displays ``OK'' is the player wins (i.e. parses correct), ``NG'' otherwise.
The game does not indicate which dependency relation was wrong to the player.

By jumping on the board, the player can have a go at the next sentence.
The software keeps tracks of the response time for which each action has taken.
Although we use a balance board as an input device, the software is implemented so that joysticks, game pad (including any censors of Nintendo Wii remote), and cursor keys on the keyboard can be replaced.

Note, whereas the direction of the arrow in the global standard is from the head to the dependent, the one of in Japanese standard is from the dependent to head. Japanese base phrase (bunsetsu) is strictly head-final dependency, which is always presented by the left arrow in the global standard and by the right arrow in the Japanese standard. There is no ambiguity to present not arrows but plain lines to indicate the dependency edges.

\section{The Design of Example Sentences \label{sec:gps}}
\ifMOVIE
\begin{figure}[tb]
\begin{center}
\includemovie[toolbar,mouse,
	text={\begin{minipage}[t]{7cm}
\begin{center}
\includegraphics[width=7cm]{fig/all.eps}
\url{http://goo.gl/EShaJP}\\
Click above to activate the movie.
\end{center}
\end{minipage}
},
]{}{}{fig/all.swf}
\end{center}
\caption{Example Sentences \label{examplesentences}}
\end{figure}
\else
\begin{figure*}[tb]
\begin{center}
\includegraphics[width=14cm]{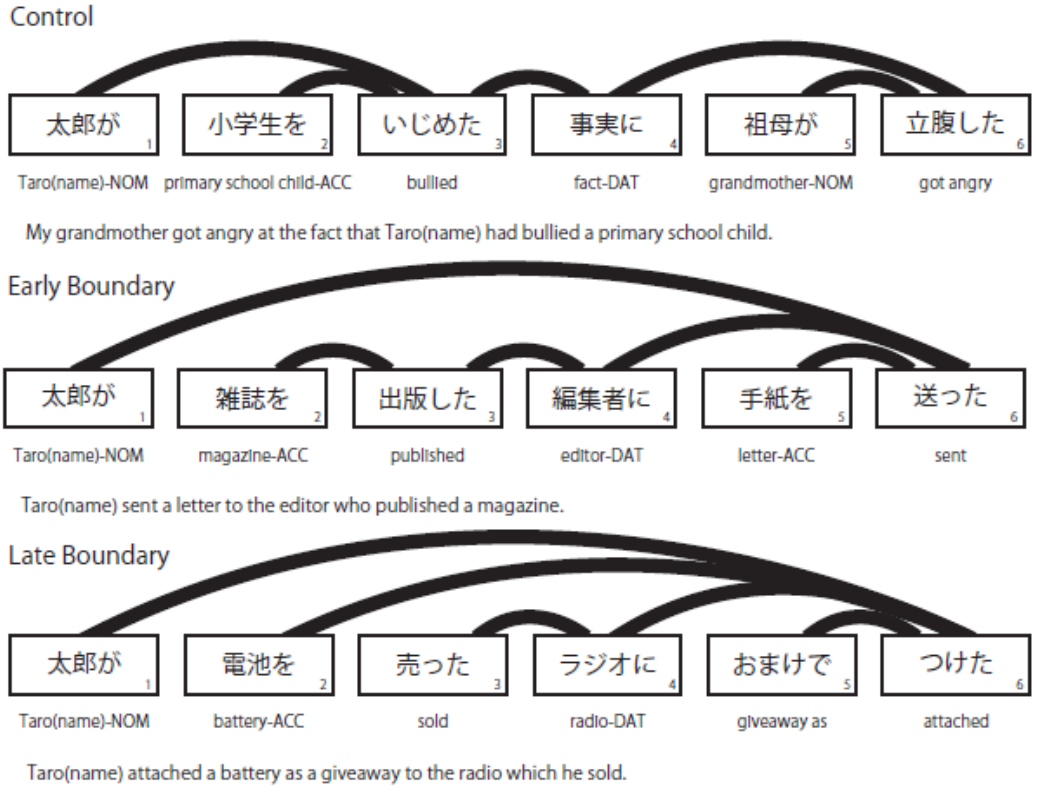}
\url{http://goo.gl/EShaJP}\\
\end{center}
\caption{Example Sentences \label{examplesentences}}
\end{figure*}
\fi

The setting of parsing difficulty is one of issues to make the game more joyful.
We focus on garden path sentences containing embedded structures presented in
Tokimoto's paper \cite{Tokimoto04} for the game.
Figure \ref{examplesentences} shows 3 types of sentences used in our experiments.

Each sentence takes 6 phrases of the form,

\begin{center}
$NP^{NOM}_1$ $NP^{ACC}_2$ $V^{PAST}_3$ $NP^{DAT}_4$ $X_5$ $V^{PAST}_6$
\end{center}

Where the 5th phrase $X_5$ is variant.
Despite case markers of the 1st $NP^{NOM}_1$, the 2nd $NP^{ACC}_2$ and the 4th $NP^{DAT}_4$ are identical,
nominatives and accusatives of the 3rd and the 6th $V^{PAST}$ vary depending on the 5th phrase.

We classify garden path sentences by the 5th phrase into 3 types.
A sentence is called Control (CTRL) if the 5th phrase is $NP^{NOM}_5$ with a nominal case marker ``ga''.
Similarly, a sentence is called Early Boundary (EB) if the 5th phrase is $NP^{ACC}_5$ with an accusative case marker ``wo''.
Otherwise, a sentence is called Late Boundary (LB).

In Figure \ref{examplesentences}, we notice how different dependency relations become among the garden path sentences.
These sentences have different dependency structures due to semantic dependency constraints of the two verbs (3rd and 6th phrases).

In the CTRL sentences, the 1st phrase $NP^{NOM}_1$ is a subject (nominative) of the 3rd phrase $V^{PAST}_3$ and the 5th phrase $NP^{NOM}_5$ is a subject of the 6th phrase $V^{PAST}_6$.
Due to the projective and head-final constraints, the 2nd $NP^{ACC}_2$ has to be a direct object (accusative) of the 3rd phrase $V^{PAST}_3$.

In the EB sentences, the 4th phrase $NP^{DAT}_4$ is a notional subject of the 3rd phrase $V^{PAST}_3$ and the 1st phrase $NP^{NOM}_1$ is a subject of the 6th phrase $V^{PAST}_6$, since a verb has only one subject phrase.
Due to the projective and head-final constraints, 
the 2nd phrase $NP^{ACC}_2$ has to be an direct object of the 3rd phrase $V^{PAST}_3$, and 
the 5th phrase $NP^{ACC}_5$ has to be an direct object of the 6th phrase $V^{PAST}_6$.

In the LB sentences, the 4th phrase $NP^{DAT}_4$ is a notional object of the 3rd phrase $V^{PAST}_3$.
Due to single accusative constraint, the 2nd $NP^{ACC}_2$ has to be a direct object of the 6th phrase $V^{PAST}_6$.
Due to the projective and head-final constraints, the 1st phrase $NP^{NOM}_1$ has to be a subject of the 6th phrase $V^{PAST}_6$.

\cite{Tokimoto04} performed psycholinguistic experiments on the garden path sentences as in Figure \ref{examplesentences} using self paced reading method and question answering method.
In his experiments, the phrases of the sentence are incrementally presented.
The reanalysis cost of the sentences for human judges has been reported as CTRL $<$ EB $<$ LB.
He also found that the reanalysis cost varied among individuals.

\section{Experiments \label{sec:experiments}}
We propose that our game application can be an interesting alternative to some what boring corpus annotation.
We use our shWiiFit Reduce Dependency Parsing to evaluate parsing difficulty of garden path sentences containing embedded structures for 12 graduate students aged 22-27.

Each experimental subject attempts a parsing game of 40 sentences in one main session.
10 sentences are the 3 types of garden path sentences originally used in Tokimoto's experiment and the remaining 30 sentences are filler sentences.
The vocabulary of the sentences is carefully controlled by the word frequency of 10-year newspaper corpus and word familiarity rating of NTT database.
Sentences are presented in the following order: 1) 5 fillers, 2) 15 fillers $+$ 5 CTRLs $+$ 5 EBs $+$ 5 LBs, 3) 5 fillers, where 1) 2) and 3) are randomized.
No sentence is duplicated and the sentence order is different for each experimental subject.

To familiarize experimental subjects with the game interface, we gave them a 5 minute instruction with a 6-page leaflet.
We used a 2-page photocopy of a 4th grade elementary school textbook to explain Japanese syntactic dependency structure.
Each experimental subject can try preliminary tests of 10 sentences and can repeat up to 3 times.
On average, 13.7 sentences are tried in the preliminary tests.
Out of 12 experimental subjects, only 1 experimental subject frequently played Nintendo Wii Fit games.

\begin{figure}
\begin{center}
\includegraphics[width=7cm]{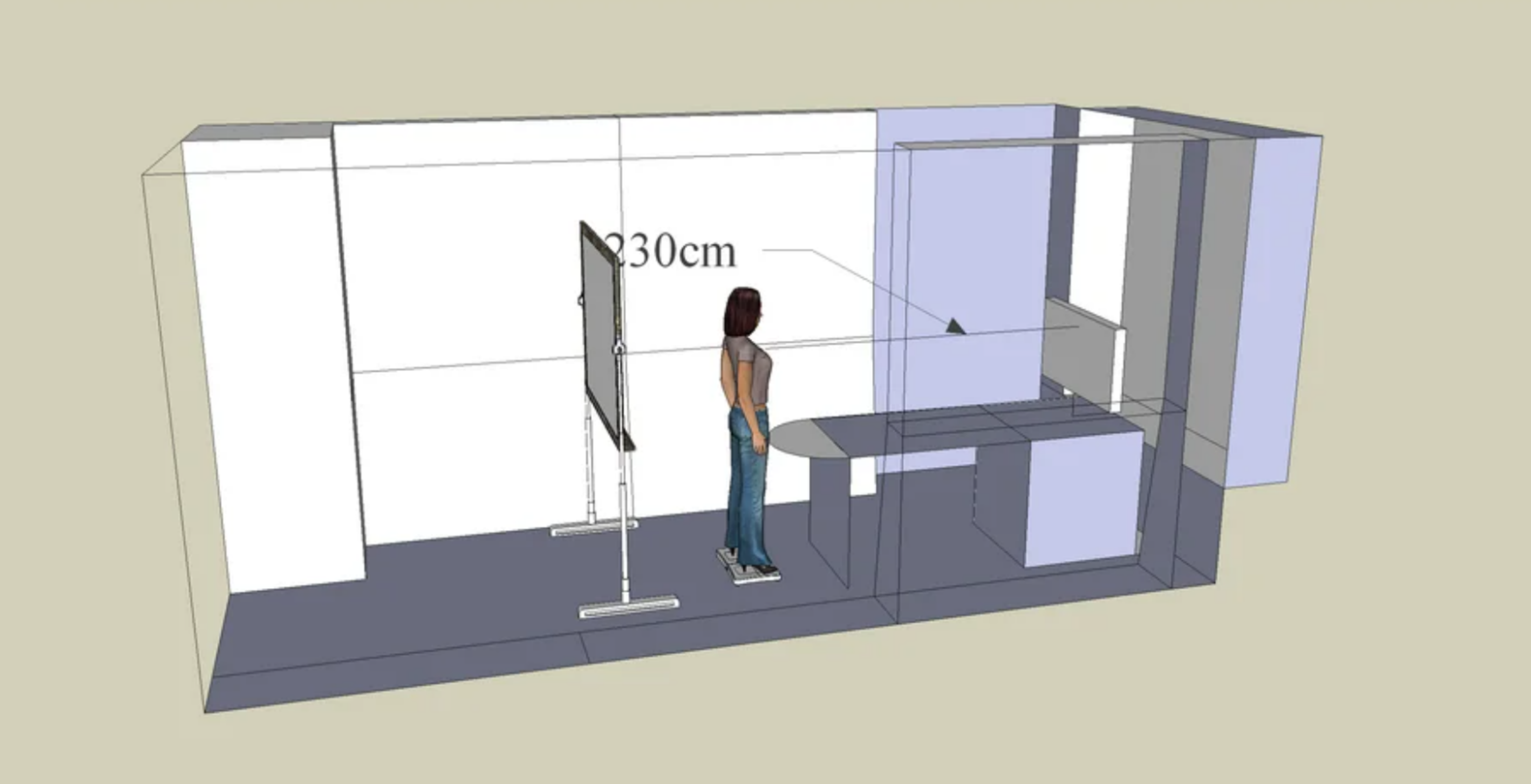} 
\url{http://goo.gl/nIe6k3}\\
\end{center}
\caption{The experimental environment \label{fig:room}}
\end{figure}

The screen is displayed in a 800x600 pixel region in the center of 42 inch display panel with 1024x768 resolution.
Though the room is not soundproofed,
a window shade of the room is pulled down, and we try to keep the room quiet.
The distance between the balance board and the display panel is 230cm.
The other specifications are indicated by measuring the 3D object in
figure \ref{fig:room} or \url{http://goo.gl/nIe6k3}.

\begin{table}[tb]
\begin{center}
\caption{Sentence accuracy and sentence response time (Human judges) \label{table:result:pacc}}
\begin{tabular}{ll|cccc} \hline\hline
   & & Filler & CTRL & EB & LB \\ \hline
acc. (\%)& ave. & 72 & 63 & 82 & 45 \\  
 & stdev. & 19.7 & 38.9 & 21.7 & 32.1 \\  \hline
s.r.r.t. & ave. & -0.12 & 0.05 & 0.12 & 0.81 \\ 
 & stdev. & 0.12 & 0.71 & 0.51 & 0.40 \\ \hline 
\end{tabular}
\end{center}
\end{table}

Table \ref{table:result:pacc} summarizes the sentence accuracy (``acc'') and sentence residual response time (``s.r.r.t.'') of the human judgments.
``acc. ave.'' is the average of sentence accuracy for Fillers, CTRLs, EBs, and LBs.
``s.r.r.t. ave.'' is an internally studentized residual of the time for the correctly parsed sessions where linear regression excluded (a) \# of morae, (b) \# of characters, (c) \# of phrases, (d) sentence order, (e) \# of ``default shift'', (f) \# of ``default reduce'', (g) \# of ``SHIFT'', (h) \# of ``REDUCE'', (i) \# of times transferring body weight between left and right.

In terms of sentence accuracy, LB is the worst and EB is better than CTRL and Filler in human judgments.
Compared with Tokimoto's experiment, our result was the same in LB sentences but different in EB sentences.
The difference came from the experiment procedure.
Our experimental subjects could read the whole sentence throughout a game.
Tokimoto showed phrases incrementally, while our experimental subjects could read the whole sentence at the beginning of a game.
In Japanese, the 1st $NP^{NOM}_1$ phrase tends to be a nominative of the last 6th $V^{PAST}_6$ phrase.
The bias helps experimental subjects to parse correctly in EB, and hinders in CTRL.

In terms of sentence residual response time, LB is longer time to be parsed correctly than the others.
There is no statistically significant difference between CTRL and EB.

Note that Table \ref{table:result:pacc} shows just a general tendency, one experimental subject parsed correctly in all EBs and LBs but make mistakes in CTRLs.
It means that the reanalysis cost and bias of human varied among individuals.

Our game application can replace the balance board with any kinds of input devices.  An effect of using the balance board is that the experimental subjects need to keep their weight centered.
The effect reduces their working memory for parsing.
In preliminary experiments with 4 subjects, the difference of accuracies between CTRL and LB with the balance board is larger than ones with simple game pads.

\section{Conclusions \label{sec:conclustions}}
We describe a game application to evaluate human dependency parsing process.
Using Nintendo Wii Balance Board as an input device, we applied the game application to evaluate difficulty to parse sentences for humans.
As far as we know, this work is the first comprehensive attempt to evaluate how humans accurately parse sentences in the linear order shift reduce manners by game application.

%Our current and future area of work is developping iOS or android version of the application.
%Crowdsourcing approach is one of solution for the objective to collect many judges by humans for sentences. 
%Another future work is to get distribution of human judge on truly ambiguous sentences which cannot be resolved by syntactic constraints. We would like to collect selectional preferences by human judges.

\section*{Acknowledgements}

We really appreciate Prof. Djam\'{e} Seddah to encourage us to publish the English version of the article.
The Japanese version is \url{ https://doi.org/10.5715/jnlp.18.351}

\bibliographystyle{acl}
\bibliography{acl2015}

\begin{thebibliography}{}

\bibitem[\protect\citename{Iwatate \bgroup et al.\egroup }2008]{Iwatate08}
Masakazu Iwatate, Masayuki Asahara, and Yuji Matsumoto.
\newblock 2008.
\newblock {Japanese Dependency Parsing Using a Tournament Model}.
\newblock In {\em Proc. of the 22nd International Conference on Computational
  Linguistics(COLING-2008)}, pages 361--368.

\bibitem[\protect\citename{Kudo and Matsumoto}2002]{Kudo02}
Taku Kudo and Yuji Matsumoto.
\newblock 2002.
\newblock {Japanese Dependency Analyisis using Cascaded Chunking}.
\newblock In {\em Proc. of the 6th Conference on Natural Language
  Learning(CoNLL-2002)}, pages 1--7.

\bibitem[\protect\citename{Nivre and Scholz}2004]{Nivre04}
Joakim Nivre and Mario Scholz.
\newblock 2004.
\newblock {Deterministic Dependency Parsing of English Text}.
\newblock In {\em Proc. of the 20th International Conference on Computational
  Linguistics(COLING-2004)}, pages 64--70.

\bibitem[\protect\citename{Sassano}2004]{Sassano04}
Manabu Sassano.
\newblock 2004.
\newblock {Linear-Time Dependency Analysis for Japanese}.
\newblock In {\em Proc. of the 20th International Conference on Computational
  Linguistics(COLING-2004)}, pages 8--14.

\bibitem[\protect\citename{Tokimoto}2004]{Tokimoto04}
Shingo Tokimoto.
\newblock 2004.
\newblock {Reanalysis Costs in Processing Japanese Sentences with Complex NP
  Structures and Homonyms: Individual Differences and Verbal Working Memory
  Constraints}.
\newblock Technical Report JCSS-TR-53, Japanese Cognitive Science Society.

\end{thebibliography}

\end{document}